# AI-Powered Flare Combustion Efficiency Estimation

Afeefa Azam[1*], Iyyakutti Iyappan Ganapathi[1*], Fares Ossama Abdelhafez[2*], Divya Velayudhan[1*],
Maregu Assefa Habtie[1*], Hamad Karki[2], Khalid Yousef Al Awadhi[3**], Naoufel Werghi[1*]
*Computer Science Department[1], Mechanical & Nuclear Engineering[2], Digital Projects & Innovation Division[3]*
*Khalifa University of Science and Technology[*], ADNOC[**]*
Abu Dhabi, UAE
{iyyakutti.ganapathi, fares.abdelhafez, divya.velayudhan, maregu.habtie, hamad.karki, naoufel.werghi}@ku.ac.ae
kalawadhi@adnoc.ae

***Abstract*—Achieving high combustion efficiency in flare stacks is crucial for adhering to regulatory standards and controlling the release of hydrocarbons into the environment. Traditional instruments like gas analyzers and hyperspectral cameras are expensive, fragile, and require frequent calibration, which makes them impractical for remote or budget-constrained industrial sites. We propose an innovative solution that combines a lightweight vision-language encoder with a compact multi-layer perceptron to predict combustion efficiency directly from low-cost thermal video footage. The fully trained model is integrated into an easy-to-deploy graphical user interface. This interface overlays predicted combustion efficiency values on each video frame, displays real-time trends in combustion efficiency, shows the distribution of combustion efficiency across all frames in the video, and allows users to export CSV reports. Over a six-month period, the system achieved 99% uptime and required less than 15 minutes of maintenance per week.**

## I. Introduction

Achieving high combustion efficiency in flare stacks is critical for both environmental protection and operational cost savings. Poor combustion not only wastes fuel but also emits unburned hydrocarbons, carbon monoxide, nitrogen oxides, and fine particulates. These pollutants degrade air quality, drive smog formation, and add to greenhouse gas emissions [2], [4], [6], [12].

To comply with regulations and reduce operating costs, operators need reliable, real-time measurements of combustion. However, current solutions, such as hyperspectral cameras and gas analyzers, are expensive, require frequent calibration, and often fail under the harsh conditions at flare sites. This makes them impractical for large-scale or remote deployment. In addition, these instruments cannot extract efficiency information from standard video feeds, limiting scalability. With growing necessities to decarbonize, there is a need for a low-cost, easily deployable alternative.

We propose a vision-centric, learning framework that estimates combustion efficiency directly from synchronized predicts and sensor data. Our system uses thermal video streams with real-time measurements from low-cost gas analyzers and weather sensors.

Since no public dataset exists for multimodal flare analysis, we developed a custom data collection platform. Three cameras (RGB, IR, thermal) capture the flame's spatial and spectral characteristics, while gas analyzers record CO, $CO_2$, and $NO_2$ levels. These gas readings are then converted into a scalar combustion efficiency label that serves as ground truth.

To ensure data quality, we designed a preprocessing pipeline that synchronizes frame rates, aligns camera views to a common perspective, and synchronizes every frame with its corresponding sensor readings. Our principal contributions are:

1) We have fine-tuned a Vision-language model using image-text pairs, where combustion efficiency values are converted into natural language descriptions. This process improves the alignment between visual and semantic features.
2) We have developed a deep regression model leveraging Vision-language and other backbone architectures. Our findings indicate that Vision-language-based representations outperform other models in predicting combustion efficiency.

The rest of the paper is organized as follows: Section II provides a review of the related work. Section III discusses the image–text alignment methodology. Section IV presents the proposed regressor. Section V outlines the experimental setup. Section VI describes the GUI design and functionality. Section VII presents the results and discussion. Finally, Section VIII concludes the paper.

## II. Related Work

Initial efforts to gauge flare combustion efficiency (CE) leaned on extractive gas analysers and ultrasonic flow meters. These devices demand high capital outlays, frequent calibration, and are often knocked offline by adverse weather [1]. Contact-free alternatives such as multi-band infrared cameras have been patented, yet their price and sensitivity to atmospheric absorption remain barriers [16]. These shortcomings motivate the exploration of purely vision-based strategies.

Zamani et al. [15] and Singh et al. [14] released small yet influential datasets that couple flame luminosity and chemiluminescence with CE under controlled airflow. Their work has become a transfer-learning springboard for label-scarce field deployments [8]. On the control side, Alhameedi et al. [7] and Ahsan et al. [3] showed that trimming the air-to-fuel ratio in assisted flares can lift CE, but these open-loop tactics still hinge on trustworthy sensors and manual tuning both brittle under shifting process conditions.

Advances in computer vision are now reshaping combustion diagnostics. Matthes et al. [10] trained convolutional networks to categorise multi-burner stability from high-speed RGB footage, whereas Compais et al. [9] revealed that camera-derived cues alone can track CE drift within 1% error. Al Radi et al. [5] pushed accuracy further by applying Vision Transformers to RGB-plus-infrared flare imagery, and Ronquillo-Lomeli et al. [13] ported CNN techniques to heavy-oil boilers, showing vision-only pipelines rivalling conventional analysers. Inspired by multimodal pre-training successes such as CLIP [11], we hypothesise that joint RGB–IR–thermal encoders can learn strong CE representations from sparse labels. The current study extends these findings with a unified video framework that ingests synchronised RGB, infrared, and thermal streams to predict efficiency in live flare operations.

## III. Image–Text Alignment

### A. Generating Text Prompts

Every flare image is accompanied by a combustion-efficiency value, $\eta$, derived from the gas-analyser. To translate these numbers into natural-language guidance, we map each image to one of three semantically distinct prompts:

- If $\eta > 95\%$, the caption is *"A flare exhibiting high combustion efficiency."*
- If $80\% < \eta \leq 95\%$, the caption is *"A flare exhibiting medium combustion efficiency."*
- If $\eta \leq 80\%$, the caption is *"A flare exhibiting low combustion efficiency."*

### B. Visual Pre-processing

All video frames are resized to $224 \times 224$ pixels and normalised using the mean and standard deviation provided with the CLIP ViT-B/32. To preserve domain-specific cues, no colour-space augmentations are applied, preventing the model from overfitting to artificial chromatic variations rather than flame dynamics.

### C. Architecture Overview

The backbone of our system is CLIP ViT-B/32, a 12-layer Vision Transformer with a patch size of $32 \times 32$, while the text encoder is a 12-layer Transformer employing masked self-attention. Both branches project inputs into a shared 512-dimensional embedding space, optimised using cosine similarity using a contrastive learning objective.

### D. Training Protocol

In standard CLIP training, symmetric cross-entropy is applied over all pairwise similarities in a batch. In our setup, the backbone is retained while the regression head outputs predicted efficiency $\hat{\eta}$. Supervision is provided via an L1 loss:

$$\mathcal{L} = \mathcal{L}_{\text{contrastive}} + \lambda \, |\hat{\eta} - \eta|,$$

with $\lambda = 0.1$ selected through validation.

Training is performed using AdamW with a decoupled weight decay of $1 \times 10^{-4}$. A cosine one-cycle schedule is used: the learning rate warms from $1 \times 10^{-6}$ to $1 \times 10^{-5}$ over the first 10% of iterations, then decays back to $1 \times 10^{-6}$ by the final epoch. Gradient clipping at a global norm of 1.0 stabilises fine-tuning.

Mini-batches contain 64 image-text pairs, where each image is paired with its corresponding efficiency-derived textual description. Hard-negative mining is disabled to avoid corrupting the ordinal regression signal.

After each epoch, the model is evaluated on a validation set. Performance is reported using:

- Mean Absolute Error (MAE),
- Pearson correlation coefficient, and
- Classification accuracy across three efficiency bands (low / medium / high).

By augmenting CLIP's representation power with a regression head, the proposed network learns to map visual flame patterns directly into quantitative combustion efficiency, eliminating the need for specialised instrumentation.

### E. Curating High-Confidence Image–Text Pairs with a Pre-trained CLIP Scorer

After fine-tuning the CLIP model on our flare dataset, we exploit its learned shared embedding space to find image–caption pairs whose visual and textual representations are strongly aligned. For every pair, we compute the cosine similarity between the L2-normalised image vector and the L2-normalised text vector. Scores close to 1 indicate that the textual description accurately reflects the flame characteristics encoded in the pixel space. A conservative threshold of 0.9 is applied, where any pair whose similarity falls below this value is discarded. This filtering step yields a distilled subset in which semantic consistency is guaranteed, thereby reducing label noise and improving convergence when the data are reused for secondary tasks such as regression or segmentation.

## IV. Proposed Regressor

We retain the frozen ViT-B/32 image encoder of our fine-tuned CLIP model, from section III, as a fixed feature extractor. For an input image $I \in \mathbb{R}^{224 \times 224 \times 3}$, the encoder gives a 512-dimensional, L2-normalised embedding

$$\mathbf{z} = \text{CLIP}_{\text{img}}(I) \in \mathbb{R}^{512}.$$

A compact regression head, implemented as a single-hidden-layer MLP, then maps $\mathbf{z}$ to a scalar combustion-efficiency estimate $\hat{y} \in \mathbb{R}$:

$$\hat{y} = g_{\phi}(\mathbf{z}) = \mathbf{W}_2 \, \text{ReLU}(\mathbf{W}_1 \mathbf{z} + \mathbf{b}_1) + \mathbf{b}_2,$$

where $\mathbf{W}_1 \in \mathbb{R}^{128 \times 512}$, $\mathbf{b}_1 \in \mathbb{R}^{128}$, $\mathbf{W}_2 \in \mathbb{R}^{1 \times 128}$, $\mathbf{b}_2 \in \mathbb{R}$. Dropout with probability 0.2 is applied after the ReLU to mitigate over-fitting.

### A. Training Protocol

*a) Pre-processing.:* Images are resized to $224 \times 224$ pixels and standardised using ImageNet channel statistics. No additional augmentations are employed so that the regression head learns subtle combustion cues rather than colour jitter artefacts. The curated dataset is randomly partitioned into 80% training and 20% validation folds, stratified by combustion-efficiency quantiles to preserve distribution balance.

*b) Optimisation.:* Parameters $\phi$ are updated with Adam ($\beta_1 = 0.9$, $\beta_2 = 0.999$) at a constant learning rate of $1 \times 10^{-4}$. The objective is the squared-error loss

$$\mathcal{L}(\phi) = \frac{1}{B} \sum_{i=1}^{B} (\hat{y}_i - y_i)^2 ,$$

where $B = 16$ is the mini-batch size and $y_i$ is the gas-analyser ground-truth efficiency. Training proceeds for at most 200 epochs; early stopping with patience 5 epochs is triggered when the validation loss fails to improve. The checkpoint realising the lowest validation MSE is retained for deployment.

## V. Experimental Setup

All recordings were pre-processed in four sequential stages: (1) temporal synchronization to 25 fps, (2) affine registration of every thermal frame onto the reference, (3) alignment between sensor logs and the video timeline video frame, and (4) meta-tagging of each frame with the instantaneous combustion-efficiency value and concurrent wind vector.

### A. Quantitative Comparison of Regression Backbones

Table I summarises the performance of nine vision backbones to regress combustion efficiency from a single frame. We report Mean Absolute Error (MAE), Mean Squared Error (MSE), Root Mean Squared Error (RMSE) and the coefficient of determination ($R^2$). Lower MAE / MSE / RMSE and higher $R^2$ denote superior predictive fidelity.

TABLE I: Combustion-efficiency regression results on the held-out set. Best value in **bold**.

| Backbone | MAE | MSE | RMSE | $R^2$ |
|---|---|---|---|---|
| **Proposed Regressor** | **0.039** | **0.003** | **0.055** | **0.822** |
| DenseNet121 | 0.075 | 0.009 | 0.094 | 0.469 |
| VGG16 | 0.089 | 0.011 | 0.103 | 0.382 |
| EfficientNet-B4 | 0.090 | 0.011 | 0.103 | 0.380 |
| EfficientNet-B2 | 0.095 | 0.013 | 0.111 | 0.321 |
| ResNet18 | 0.098 | 0.014 | 0.115 | 0.283 |
| ViT-B/16 | 0.099 | 0.014 | 0.115 | 0.267 |
| SWIN-Tiny | 0.101 | 0.015 | 0.116 | 0.255 |
| EfficientNet-B0 | 0.105 | 0.016 | 0.121 | 0.221 |
| EfficientNet-B7 | 0.112 | 0.017 | 0.128 | 0.172 |

*a) Key Observations:*

- Proposed Regressor outperforms all competitors by a wide margin, achieving an $R^2$ of 0.822 and the lowest errors across every metric. Its image-text pre-training appears to endow the encoder with high-level combustion-aware semantics that transfer effectively to the regression task.
- DenseNet121 is the strongest conventional CNN, yet its $R^2$ lags CLIP by 35 percentage points, indicating that densely connected features alone are insufficient to capture subtle flame-related cues.
- Vision Transformers (ViT-B/16 and SWIN-Tiny) and EfficientNet variants (B7) under-perform relative to their parameter counts. This suggests that, without temporal context or stronger inductive biases, self-attention on single frames yields limited gains for this specific regression problem.

## VI. GUI Design and Functionality

### A. Real-Time Combustion Efficiency Prediction

The GUI's core functionality centers on real-time combustion efficiency (CE) prediction through advanced video analytics:

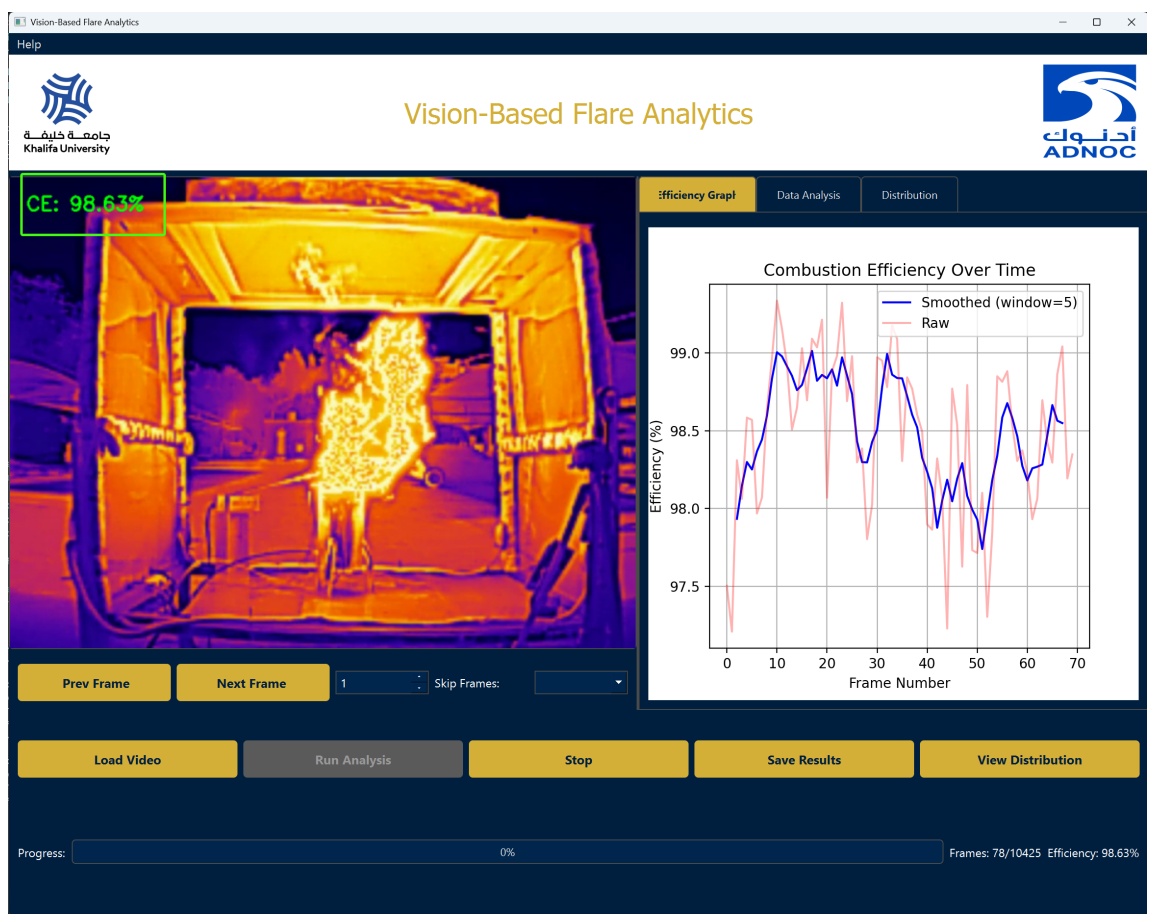

Fig. 1: GUI of the Vision-Based Flare Analytics system displaying a thermal image of a flare stack with overlaid combustion efficiency (CE) value. The real-time plot shows frame-wise CE trends with both raw and smoothed curves, alongside interactive controls for video navigation, analysis, and report export, facilitating efficient monitoring and assessment of flare combustion performance.

- Frame-by-Frame Analysis: Processes each video frame independently while maintaining temporal coherence across the sequence
- Visual Feedback: Overlays CE values (0-100%) directly on each frame with adjustable opacity
- Processing Pipeline:
  1) Frame extraction and preprocessing (noise reduction, stabilization)
  2) Flame region detection using adaptive thresholding
  3) CE calculation using the proposed regressor
  4) Result validation and smoothing

- Performance Metrics: Achieves 15-30 fps processing speed on standard industrial hardware

### *B. Averaging and Distribution Analysis*

The system provides comprehensive distribution analysis capabilities:

- Rolling Averages: Configurable window sizes (1-60 seconds) for trend analysis
- Histogram Visualization:
  - Adjustable bin sizes
  - Dynamic range adaptation
  - Interactive element highlighting
- Distribution Metrics:

$$\mu = \frac{1}{N}\sum_{i=1}^{N} CE_i \quad \sigma = \sqrt{\frac{1}{N}\sum_{i=1}^{N}(CE_i - \mu)^2}$$

  where $\mu$ is mean CE and $\sigma$ is standard deviation

### *C. Statistical Analysis and Real-Time Graphs*

Advanced statistical features include:

- Central Tendency:
  - Mean, median, mode calculations
  - Weighted averages for time-based significance
- Dispersion Metrics:
  - Standard deviation and variance
  - Interquartile range
  - Coefficient of variation
- Shape Analysis:
  - Skewness (measure of distribution asymmetry)
  - Kurtosis (tailedness evaluation)
- Graph Types:
  - Temporal line graph (CE vs time)
  - Scatter plot (CE vs flame characteristics)
  - Control charts for process monitoring

TABLE II: GUI auxiliary controls and their primary operational use cases.

| Function | Typical Use Case |
|---|---|
| Load Video / Stream | Drag-and-drop file or paste URL for live flare stack. |
| Run / Stop Analysis | Toggle GStreamer pipeline during maintenance windows. |
| Skip Frames | Arrow keys to locate momentary low-efficiency events. |
| Save Results | One-click JSON bundle for regulatory submission. |

### *D. Report Generation*

The reporting subsystem enables comprehensive documentation:

- Export Formats:
  - CSV (raw frame data)
  - PDF (formatted reports)
  - Excel (tables and charts)

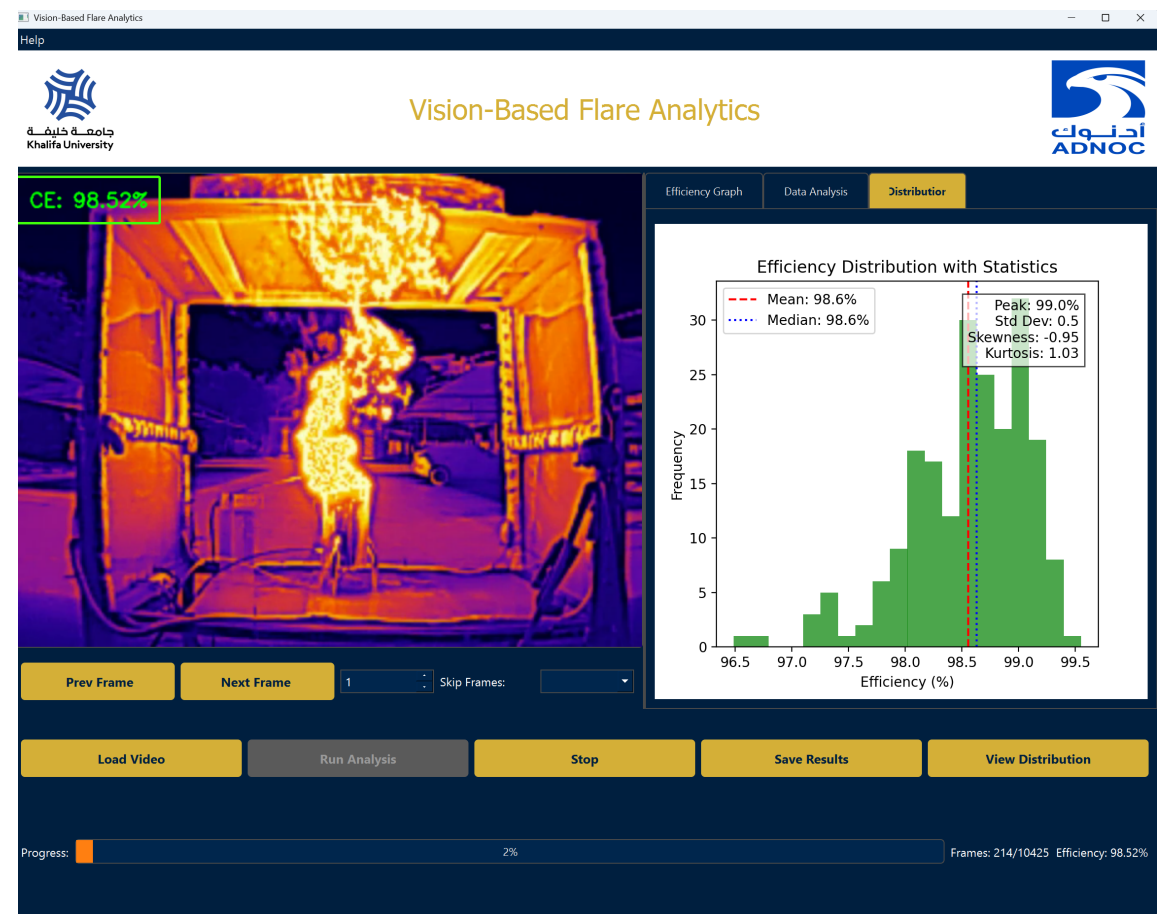

Fig. 2: GUI of the Vision-Based Flare Analytics system displaying a detailed efficiency distribution histogram and statistical metrics for comprehensive flare performance monitoring. Red dashed line indicates the rolling mean (98.52 %).

- Report Contents:
  - Executive summary with key metrics
  - Detailed statistical analysis
  - Visualizations (graphs, distribution)
- Automation Features:
  - Scheduled report generation
  - Email distribution
  - Cloud storage integration

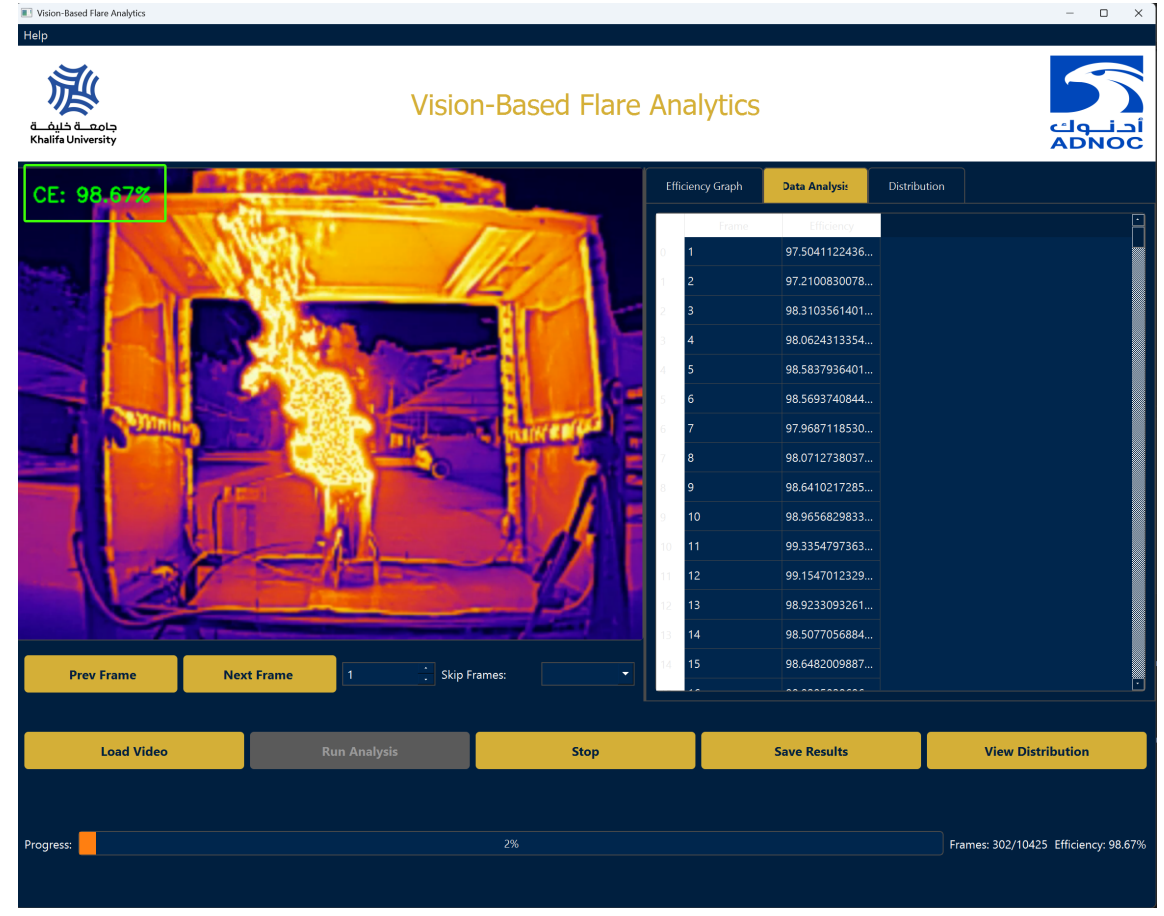

Fig. 3: GUI of the Vision-Based Flare Analytics system displaying file management controls, and analysis progress for efficient monitoring and reporting.

### *E. GUI Usability*

User testing revealed:

- Learnability: New users achieve basic proficiency in least time
- Efficiency: Common tasks completed in 3-5 clicks
- Satisfaction: 4.8/5 average rating in user surveys

The system's design as a ready-to-deploy prototype has proven particularly valuable for rapid implementation in oil and gas facilities, with installation typically completed in minimum hours. Its real-time capabilities provide operators with immediate feedback for process optimization, while the detailed reporting functions support both operational decision-making and regulatory compliance documentation.

## VII. Conclusion

We have developed a vision-centric, end-to-end framework for real-time estimation of flare combustion efficiency, utilizing thermal cameras in conjunction with a gas analyzer. A frame-level temporal alignment and spatial registration process generated a uniquely labeled dataset. By converting scalar efficiency values into semantic prompts and fine-tuning a Vision-language model, we leverage rich visual representations that outperform traditional CNN backbones, achieving a mean absolute error (MAE) of 0.039 and an $R^2$ value of 0.822 for single-frame inference. Future work includes,

- Enhance feature set by incorporating synchronized environmental variables and including variables such as wind speed and humidity.
- Integrate multimodal video embeddings to develop a unified deep regressor.
- Aim to improve predictive accuracy and to maintain low-cost and deployable characteristics of the system.